\documentclass[journal]{IEEEtran}
\usepackage[utf8]{inputenc}

\usepackage{amsmath}
\usepackage{amssymb}
\usepackage{dsfont}

\usepackage{tikz}
\usetikzlibrary{matrix,calc}

\usepackage{graphicx}
\graphicspath{{Graphics/}}
\usepackage{subcaption}

\usepackage{multirow}

\newcommand\independent{\protect\mathpalette{\protect\independenT}{\perp}}
\def\independenT#1#2{\mathrel{\rlap{$#1#2$}\mkern2mu{#1#2}}}

\newcommand{\threepartdef}[6]
{
	\left\{
		\begin{array}{lll}
			#1 & \mbox{if } #2 \\
			#3 & \mbox{if } #4 \\
			#5 & \mbox{if } #6
		\end{array}
	\right.
}

\begin{document}

\title{Learning to quantify emphysema extent: What labels do we need?}

\author{Silas Nyboe Ørting, Jens Petersen, Laura H. Thomsen, Mathilde M. W. Wille and Marleen de Bruijne~\IEEEmembership{Member,~IEEE,}%
\thanks{This study was financially supported by the Danish Council for Independent Research (DFF) and  the Netherlands Organization for Scientific Research (NWO). The sponsors had no involvement in the work.}%
\thanks{S. Ørting and J. Petersen are with Department of Computer Science, University of Copenhagen, Copenhagen, Denmark}%
\thanks{L. Thomsen is with Department of Internal Medicine, Hvidovre Hospital, Copenhagen Denmark}%
\thanks{M. M. W. Wille is with Department of Diagnostic Imaging, Bispebjerg Hospital, Copenhagen, Denmark}%
\thanks{M. de Bruijne is with Department of Computer Science, University of Copenhagen, Copenhagen, Denmark and Biomedical Imaging Group Rotterdam, Departments of Radiology and Medical Informatics, Erasmus MC - University Medical Center Rotterdam, The Netherlands }
\thanks{Manuscript received September 7th, 2018}}
\maketitle

\begin{abstract}
Accurate assessment of pulmonary emphysema is crucial to assess disease severity and subtype, to monitor disease progression and to predict lung cancer risk. However, visual assessment is time-consuming and subject to substantial inter-rater variability and standard densitometry approaches to quantify emphysema remain inferior to visual scoring. 
We explore if machine learning methods that learn from a large dataset of visually assessed CT scans can provide accurate estimates of emphysema extent. We further investigate if machine learning algorithms that learn from   a scoring of emphysema  extent can outperform algorithms that learn only from a  scoring of emphysema presence. 
We compare four Multiple Instance Learning classifiers that are  trained on emphysema presence labels, and five Learning with Label Proportions classifiers that are trained on emphysema extent labels. 
We evaluate performance on 600 low-dose CT scans from the Danish Lung Cancer Screening Trial and find that learning from emphysema presence labels, which are much easier to obtain, gives equally good performance to learning from emphysema extent labels.
The best classifiers achieve intra-class correlation coefficients around 0.90 and average overall agreement with raters of 78\% and 79\% on six emphysema extent classes versus inter-rater agreement of 83\%.
\end{abstract}

\section{Introduction}
\label{sec:introduction}
\IEEEPARstart{E}{mphysema} is a lung pathology characterized by destruction of lung tissue and enlargement of airspaces in the lung, causing shortness of breath. It is a main component of chronic obstructive pulmonary disease (COPD), a leading cause of mortality and morbidity world-wide \cite{GOLD}. Emphysema can be assessed on chest CT scans and its extent quantified by densitometry, where the amount of tissue affected by emphysema is estimated by measuring the percentage of lung volume with attenuation below a specific threshold. Although densitometry is simple and provides a single interpretable measurement of emphysema extent, it is also highly dependent on scanner hardware, reconstruction parameters \cite{sieren2012reference} and software used for analysis \cite{wielputz2014variation}.

An alternative to densitometry is visual assessment that can quantify extent and characterize emphysema subtype. The COPDGene CT Workshop Group \cite{barr2012a} proposed a standard for visual assessment of COPD based on the characterization of emphysema appearance from the Fleischner society \cite{hansell2008fleischner}. A slightly modified version of the standard was used for visual assessment in the Danish Lung Cancer Screening Trial (DLCST), where it was shown to be predictive of lung cancer \cite{wille2015visual}. A similar classification scheme defined in \cite{lynch2015ct} was used in \cite{lynch2018ct} where it was shown that visual presence and severity of emphysema is associated with increased mortality independent of densitometric measures of emphysema severity. The downside of visual assessment is that it is time-consuming and subject to inter-rater variability \cite{barr2012a,wille2014emphysema}. 

Automated approaches based on the appearance of emphysema could provide fast and reproducible assessment of emphysema extent, location and sub-type, thus combining the superior disease characterization of visual assessment with the ease of densitometry. For instance \cite{wiemker2017automated} has shown that a shape-model of bullae-like structures can be used for emphysema detection. We have previously used machine learning algorithms based on texture features to predict regional emphysema presence \cite{orting2018detecting} and emphysema extent \cite{orting2016quantifying}. Other learning based approaches have focused on discovery of emphysema patterns using supervised \cite{castaldi2013distinct} and unsupervised \cite{binder2016unsupervised,yang2017unsupervised} learning, COPD detection and staging \cite{sorensen2012texture,cheplygina2014classification} and emphysema detection in the more general context of interstitial lung disease classification \cite{wang2017multi,gao2018holistic}.

Multiple Instance Learning (MIL) has been used with success in a number of the prior works on emphysema and COPD detection \cite{orting2018detecting,sorensen2012texture,cheplygina2014classification} and for many related medical image analysis tasks as reviewed in \cite{quellec2017multiple}. MIL is a learning setting where the objects of interest are represented by a collection of samples. Each collection has a binary label and the goal is to learn which samples in a collection are ``responsible'' for the label. 
MIL has been very succesful at detecting presence of abnormalities. However, visual assessment systems for lung disease, such as those developed for COPD \cite{barr2012a}, give estimates of affected lung tissue that is better captured by proportion labels. Label Proportions Learning (LLP) is the natural extension of MIL to cases where labels are proportions, but despite the success of MIL, LLP has seen almost no usage in medical imaging.

In this work we present the largest comparison yet of machine learning methods for assessing emphysema extent, extending our previous work on emphysema presence prediction \cite{orting2018detecting}, where a MIL method was used for regional emphysema detection, and our work on extent prediction \cite{orting2016quantifying}, where the LLP method Cluster Model Selection was used for regional emphysema extent prediction. We compare four MIL methods, of which three have not been used for emphysema detection before, and five LLP methods, of which four have not been used for emphysema detection or in medical imaging before. We investigate if learning from emphysema extent labels improves performance over learning from emphysema presence labels. Knowing what can be achieved by learning from labels of different quality and cost is paramount for cost-effective development and application of machine learning methods for clinical decision making.

\section{Materials and methods}
\label{sec:materials-methods}
We view emphysema extent prediction as a bag learning problem. Bag learning is a machine learning setting where we are given a set of instances, a partition of the instances into bags and a labeling of the bags. The objective is to learn to predict both instance and bag labels for unseen data. In this work we view a region of the lung as a bag and patches sampled from the region as instances. The bag labels are regional emphysema extent scores, corresponding to estimated percentage of affected lung volume, and we wish to predict which patches contain emphysema, as well as the extent of emphysema in the region. 
Representing a scan as a set of patches provides a representation of local patterns in the lungs. By controlling the patch size we can focus on the scale at which patterns are expected to be distinct.

More formally, let $\mathcal{X}$ be an instance space, $\mathcal{Y}$ an instance label space, $\mathcal{Z}$ a bag label space and $\mathbf{b} = (\mathbf{x} \subseteq \mathcal{X}, z \in \mathcal{Z})$ a labeled bag of instances. We use superscripts to refer to the label ($\mathbf{b}^z$), instances ($\mathbf{b}^{\mathbf{x}}$) and instance labels ($\mathbf{b}^{\mathbf{y}}$) associated with a bag $\mathbf{b}$. For a set of $m$ bags $\mathbf{B} = \{\mathbf{b}_1, \mathbf{b}_2, \dots, \mathbf{b}_m\}$, $\mathbf{b}^{\mathbf{x}}_i$ are the instances in the $i$'th bag and $\mathbf{b}^{\mathbf{x}}_{ij}$ is the $j$'th instance in the $i$'th bag. We define the learning problem as
\begin{equation}
  \label{eq:bag-learning}
  \arg\max\limits_{\mathbf{Y},h,\Theta} \mathrm{P}( \mathbf{Y}, h, \Theta | \mathbf{B} ), 
\end{equation}
where $\mathbf{Y} = \cup_{i=1}^m \mathbf{b}_i^{\mathbf{y}}$ is a labeling of instances, $\Theta : \mathcal{Y} \mapsto \mathcal{Z}$ is a bag labeling function relating $\mathbf{b}_i^\mathbf{y}$ to $\mathbf{b}_i^z$ and $h : \mathcal{X} \mapsto \mathcal{Y}$ is a hypothesis relating the instances $\mathbf{b}_i^\mathbf{x}$ to the corresponding instance labels $\mathbf{b}_i^\mathbf{y}$, i.e. $h$ is a method for predicting $\mathbf{b}_i^\mathbf{y}$ from $\mathbf{b}_i^\mathbf{x}$.

Two well known bag learning settings are multiple instance learning (MIL) and learning with label proportions (LLP). In the standard MIL setting bag labels are binary, instance labels are binary and bag labels are related to instance labels by the max rule, i.e. a bag is positive if at least one instance is positive
\begin{equation}
  \label{eq:max-comb-rule}
  \mathbf{b}^z_i = \Theta_{\max}(\mathbf{b}^{\mathbf{y}}_i) = \max_j \mathbf{b}^\mathbf{y}_{ij}.
\end{equation}
This MIL setting is powerful because it allow us to learn about instance labels when only little information about the relation between instance and bag labels is available. A potential issue with the max rule is that it focuses on the single most discriminative instance. This could lead to a situation with good bag-level detection but poor localization and extent prediction. Including information about the proportion of positive instances could improve localization and extent prediction. In the standard LLP setting, bag labels are proportions, instance labels are binary and bag labels are related to instance labels by the mean rule, i.e. the bag label is the proportion of positive instances
\begin{align}
  \label{eq:mean-bag-label}
  \mathbf{b}^z_i = \Theta_{\mathrm{mean}}(\mathbf{b}_i^{\mathbf{y}}) = \frac{1}{|\mathbf{b}_i|} \sum\limits_j^{|\mathbf{b}_i|} \mathbf{b}^{\mathbf{y}}_{ij}.
\end{align}
Although MIL methods require binary labels for training, i.e. $\Theta : \mathcal{Y} \mapsto \{0,1\}$, we can use $\Theta_{\mathrm{mean}}$ at test time to obtain proportion estimates of emphysema extent.
\subsection{Methods}
\label{sec:methods}
We compared four MIL methods (logistic, SVM, $mi$-logistic, $mi$-SVM) and five LLP methods (beta, Cluster Model Selection, $\propto$-SVM, $\propto$-logistic, Laplacian Mean Map). The methods can be grouped into three distinct strategies used to solve the bag learning problem: the simple strategy, the relabeling strategy and the mean strategy. Some methods have previously been successfully applied to emphysema and COPD prediction, logistic, SVM and $mi$-SVM in \cite{cheplygina2014classification,orting2018detecting} and Cluster Model Selection in \cite{orting2016quantifying}. The LLP methods, $\propto$-SVM \cite{yu2013proptosvm} and Laplacian Mean Map \cite{patrini2014almost}, have been shown to perform well on a variety of datasets. The beta method \cite{ferrari2004beta} can be seen as an LLP version of logistic and the $mi$-logistic and $\propto$-logistic methods are logistic regression versions of their SVM counterparts.

\paragraph{Simple strategy}
\label{sec:simple}
In the simple strategy the bag learning problem is solved by ignoring intra-bag dependencies. We assign each instance the label of the bag it came from, i.e $\mathbf{b^y}_{ij} = \mathbf{b}^z_i$, and train a standard supervised method on the instance labels. Labels for unseen bags are predicted by predicting instance labels and using $\Theta_{\mathrm{mean}}$ to derive a bag label. The learning problem now becomes
\begin{equation}
  \label{eq:obj}
  \arg\max\limits_{\phi} \mathrm{P}( h_\phi | \mathbf{Y}, \mathbf{X}),
\end{equation}
where $\mathbf{X} = \cup_{i=1}^m \mathbf{b^x}_i$ is the set of instances and $h$ is a model parameterized by $\phi$. We consider two simple MIL models, logistic regression (log) and a support vector machine (svm); and one simple LLP model, beta regression \cite{ferrari2004beta} (beta). Beta regression is a generalized linear model where the outcome $\mathbf{Y}$ follows a beta distribution allowing us to perform regression with proportion outcomes. Note that bag labels are only used for the initial instance labeling, so $\Theta$ plays no role in the simple strategy.

\paragraph{Relabeling strategy}
\label{sec:relabeling-strategy}
In the relabeling strategy the bag learning problem is solved by splitting it into two sub problems that are solved separately, a standard learning problem (\ref{eq:relabel-1}) and an instance labeling problem (\ref{eq:relabel-2}),
\begin{align}
  &\arg\max\limits_{\phi} \mathrm{P}( h_\phi | \mathbf{Y}, \mathbf{X} ) \label{eq:relabel-1}\\
  &\arg\max\limits_{\mathbf{Y}} \mathrm{P}( \mathbf{Y} | h_\phi, \Theta,  \mathbf{Z} ), \label{eq:relabel-2}
\end{align}
where $\mathbf{Z} = \cup_{i=1}^m \{\mathbf{b}^z_i\}$ is the set of bag labels and $\Theta = \Theta_{\max}$ for MIL and $\Theta = \Theta_{\mathrm{mean}}$ for LLP. The two sub problems are iterated until convergence, with the result of (\ref{eq:relabel-1}) being used for (\ref{eq:relabel-2}) and the result of (\ref{eq:relabel-2}) being used for (\ref{eq:relabel-1}). 
We consider two relabeling MIL methods, $mi$-SVM \cite{andrews2003support} (misvm) and $mi$-logistic (milog); and three relabeling LLP methods, $\propto$-SVM \cite{yu2013proptosvm} (psvm), $\propto$-logistic (plog) and Cluster Model Selection \cite{stolpe2011learning} (cms). The methods milog and plog have not previously been published, they are however very similar to their svm counterparts and we do not include the derivation here. Details can be found in Appendix~\ref{app:methods}. The cms algorithm differs from the other relabeling methods in that it solves (\ref{eq:relabel-1}) by unsupervised clustering. We use a version of cms previously described in \cite{orting2016quantifying}.

\paragraph{Mean strategy}
\label{sec:mean-strategy}
In the mean strategy the bag learning problem is solved by replacing the direct dependence on instance labels with a dependence on a mean statistic $\boldsymbol{\mu}$ calculated over all instances
\begin{equation}
  \arg\max\limits_{\phi} \mathrm{P}( h_\phi | \boldsymbol{\mu}, \mathbf{X} ).
\end{equation}
$\boldsymbol{\mu}$ is defined as
\begin{equation}
  \label{eq:mean-operator}
  \boldsymbol{\mu} = \frac{1}{n}\sum\limits_{i} \mathbf{Y}_i\mathbf{X}_i 
\end{equation}
where $\mathbf{Y}_i \in \{-1,1\}$ and $n$ is the number of instances. Knowing $\boldsymbol{\mu}$ allow us to minimize the expected risk of a large class of loss functions. However, since the instance labels $\mathbf{Y}$ are still unknown $\boldsymbol{\mu}$ must be estimated. The basic idea for the mean strategy is to express $\boldsymbol{\mu}$ in terms of bag-wise averages and solve for these bag-wise averages
\begin{align}
  \boldsymbol{\mu} &= \sum\limits_{i=1}^m \frac{|\mathbf{b}_i|}{n} \boldsymbol{\mu}_i     \label{eq:mean-operator-2}\\
  \boldsymbol{\mu}_i &= \mathbf{b}^z_i \boldsymbol{\mu}^+_i - (1 - \mathbf{b}^z_i)\boldsymbol{\mu}^-_i   \label{eq:mean-operator-3}
\end{align}
where $|\mathbf{b}_i|$ the number of instances in bag $i$ and $\boldsymbol{\mu}_i, \boldsymbol{\mu}^+_i, \boldsymbol{\mu}^-_i$, are the unknown mean instance, mean positive instance and mean negative instance of bag $i$, respectively.
Equation~(\ref{eq:mean-operator-3}) yields an underdetermined system of equations. We consider a single mean LLP method, Laplacian Mean Map \cite{patrini2014almost} (lmm), that solves the system of equations by regularizing with a bag similarity term. We refer to \cite{patrini2014almost} for further details.

\subsection{Measures}
\label{sec:measures}
We measure agreement in the following way. Let $n_k$ be the number of ratings for case $k$ and $n_{c,k}$ the number of times label $c$ is assigned to case $k$. Agreement on label $c$ over all cases is defined as
\begin{equation}
  \label{eq:specific-agreement}
  \frac{ \sum\limits_{k} n_{c,k} \left(n_{c,k} - 1 \right) }{ \sum\limits_{k} n_{c,k} \left( n_k - 1 \right) }.
\end{equation}
Overall agreement across labels is defined as
\begin{equation}
  \label{eq:overall-agreement}
  \frac{ \sum\limits_{c,k} n_{c,k} \left(n_{c,k} - 1 \right) }{ \sum\limits_{k} n_k \left( n_k - 1 \right) }.
\end{equation}
When all cases have two ratings Equation~\ref{eq:specific-agreement} corresponds to the Jaccard similarity and Equation~\ref{eq:overall-agreement} corresponds to multi-class accuracy. For multiple raters these measures ensure that partial agreement, e.g. two out of three, is counted appropriately. 
We measure prevalence of label $c$ as the proportion of times a case is assigned label $c$ out of all assignments.
\begin{equation}
  \frac{ \sum\limits_{k} n_{c,k} }{ \sum\limits_{c,k} n_{c,k} }
\end{equation}

\subsection{Data}
\label{sec:data}
Examples of the appearance of emphysema in CT scans are provided in Appendix~\ref{app:emphysema}.

\subsubsection{Study population, CT scanning \& visual assessment}
\label{sec:dlcst}
We used data collected in the Danish Lung Cancer Screening Trial (DLCST) \cite{pedersen2009dlcst}. The screening arm of the study enrolled 2052 participants for annual low dose CT screening. Scan parameters are reproduced below verbatim from \cite{pedersen2009dlcst}.
\begin{quotation}
  All CT scans of the study were performed on a MDCT scanner (16 rows Philips Mx 8000, Philips Medical Systems, Eindhoven, The Netherlands). Scans were performed supine after full inspiration with caudocranial scan direction including the entire ribcage and upper abdomen with a low dose technique, 120kV and 40 mAs. Scans were performed with spiral data acquisition with the following acquisition parameters: Section collimation 16 $\times$ 0.75 mm, pitch 1.5, rotation time 0.5 second.
\end{quotation}
We used a 1mm reconstruction with pixel size of 0.78mm $\times$ 0.78mm.

We obtained visual assessment of emphysema from \cite{wille2014emphysema}, where screening participants with at least two CT scans were selected for visual assessment (n=1990). The visual assessment used a slight modification of the assessment sheets from \cite{barr2012a}. Baseline and final followup scan was assessed by two experts. Emphysema extent was assessed for the top, middle and lower regions of each lung. The regions were defined as above carina, between carina and lower pulmonary vein, and below lower pulmonary vein. Each region was assigned a score of 0\%, 1-5\%, 6-25\%, 26-50\%, 51-75\% or 76-100\% indicating the extent of emphysema in the region.

In general, prevalence was highest and rater agreement best in the upper regions. Prevalence and agreement for the upper right region are summarized in Table~\ref{tab:rater-agreement-ru}.
Prevalence for emphysema extent above $26\%$ is low ($\approx 36$ of 1200 subjects). Agreement on the five categories indicating emphysema presence was around 50\%. Using only two categories (0\%, $\ge 1\%$) improves agreement to 82\% on the emphysema category. Although the original six categories provide more information than presence/absence labels, they are noisier and likely harder to learn from.

\begin{table}[!t]
  \resizebox{\columnwidth}{!}{
    \begin{tabular}{rccrcc}
    \multicolumn{3}{c}{All}               & \multicolumn{3}{c}{Presence}  \\ 
    Extent     & Agreement   & Prev       & Extent                     & Agreement                    & Prev\\
    \hline
    0\%        & 94 (93--95) & 75.2       & 0\%                        & 94  (93--95)                 & 75.2   \\ 
    1-5\%      & 54 (47--60) & 14.7       & \multirow{5}{*}{$\ge 1\%$} & \multirow{5}{*}{81 (78--85)} & \multirow{5}{*}{24.8}\\ 
    6-25\%     & 44 (34--53) &  7.0       &                            &                              & \\ 
    26-50\%    & 45 (26--61) &  2.0       &                            &                              & \\
    51-75\%    & 57 (26--80) &  0.9       &                            &                              & \\ 
    $\ge 76\%$ & 67 (00--99) &  0.1       &                            &                              & \\ 
    \hline
    \hline
    Overall    & 83 (81--85) &            &                            & 91 (89--92)                  & \\ 
    \end{tabular}
  }
  \caption{Agreement and mean prevalence in the upper right region of the training data. Numbers are percentages. First three columns are for  all six categories, last three columns are for presence/absence. 95\% confidence intervals for agreement estimated by bootstrapping are given in parenthesis.}
  \label{tab:rater-agreement-ru}
\end{table}

\subsubsection{Patches}
\label{sec:data-patches}
We represented a lung region as a collection of 3D patches sampled from the region. Sampling was done by choosing patch center locations uniformly at random within the region. We used a fixed patch size of approximately $11\mathrm{mm}^3$ to match the size of the secondary lobule \cite{hansell2008fleischner} and allowed overlapping patches. For each patch we extracted a set of multi-scale filter responses and used equalized histograms of the filter responses as the final representation of the patch. The filters used were Gaussian blur, gradient magnitude, eigenvalues of the Hessian, Laplacian of Gaussian, Gaussian curvature and the Frobenius norm of the Hessian. All filters were calculated at scales 1mm, 2mm and 4mm. The filters and the patch sampling strategy have previously been used successfully for COPD texture analysis in \cite{sorensen2012texture}.

\section{Experiments and results}
\label{sec:experiments}
We created a set of 1800 bags by sampling patches from the upper right region of 1800 subjects, such that each bag corresponds to one unique subject. We chose the upper right region because it has the highest prevalence and agreement. Results in \cite{orting2018detecting} indicate that although absolute performance decreases when training on regions with lower prevalence and agreement, this decrease is relatively smaller than the decrease in rater agreement and prevalence.

Each bag contained 100 patches from a single subject. The bags were split into three non-overlapping datasets of 400 training and 200 test bags. Each experiment was run on all three datasets. In each split, we used two-fold cross validation on the training bags for parameter tuning. The three separate sets of classifiers were finally trained on all 400 training bags and performance estimated on the corresponding 200 test bags. 

All classifiers provide posterior instance label probabilities which were converted to binary predictions using a classifier specific instance threshold fitted on the training bags. Parameters are summarized in Appendix~\ref{app:parameters}.

To train and evaluate we derived point estimates of emphysema extent by converting visually assessed extent intervals to interval midpoints and taking the mean over both raters. As an example, for a region with ratings 6-25\% and 1-5\%, the ratings are converted to 15.5\% and 3\% and combined into 9.25\%. The point estimates where used directly for training LLP classifiers and thresholded at zero to obtain binary labels for training MIL classifiers.

\subsection{Extent prediction accuracy}
\label{sec:extent}
\begin{figure*}[!t]
  \centering
  \includegraphics[width=0.354\textwidth,trim={00 00 30 25},clip=true]{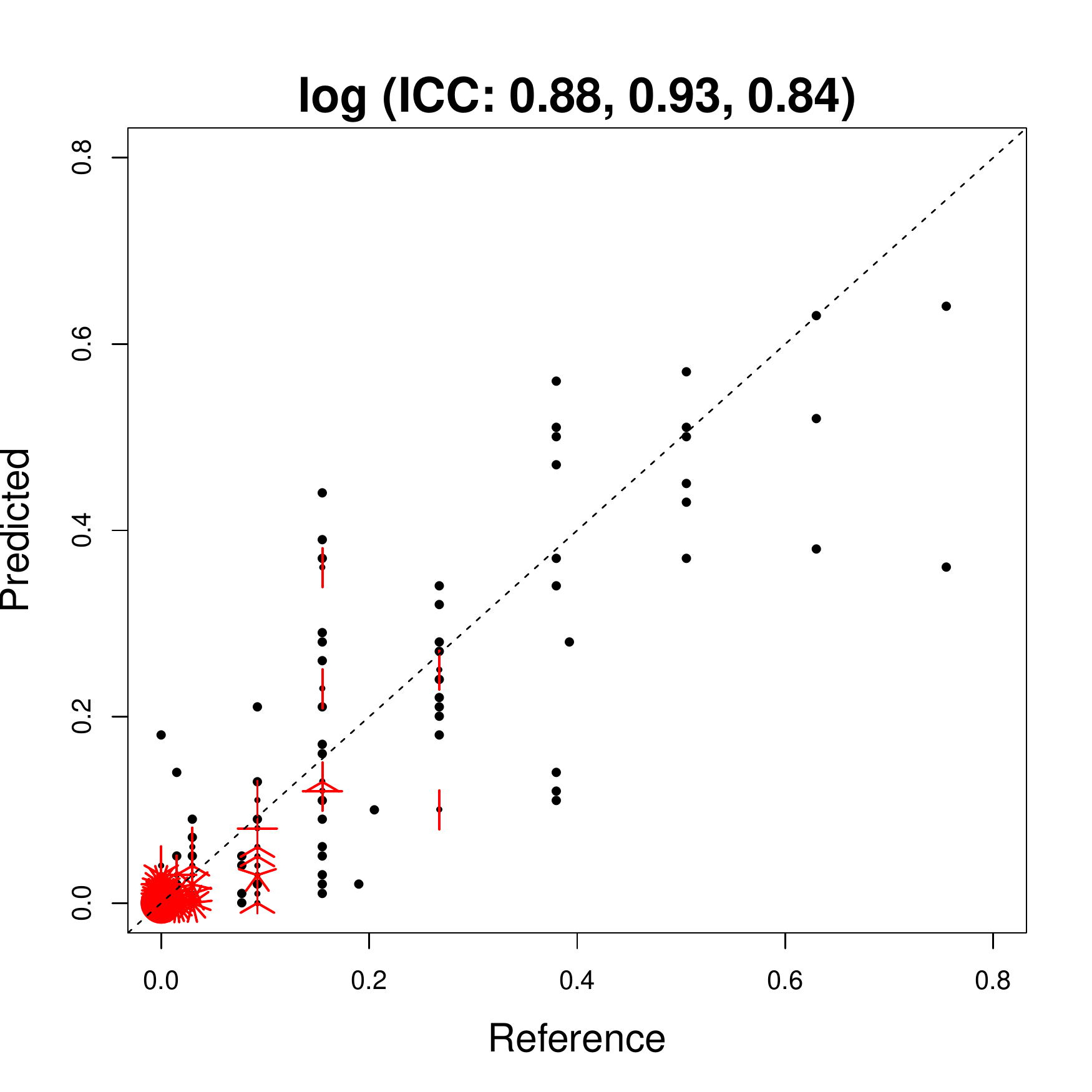}
  \includegraphics[width=0.313\textwidth,trim={55 00 30 25},clip=true]{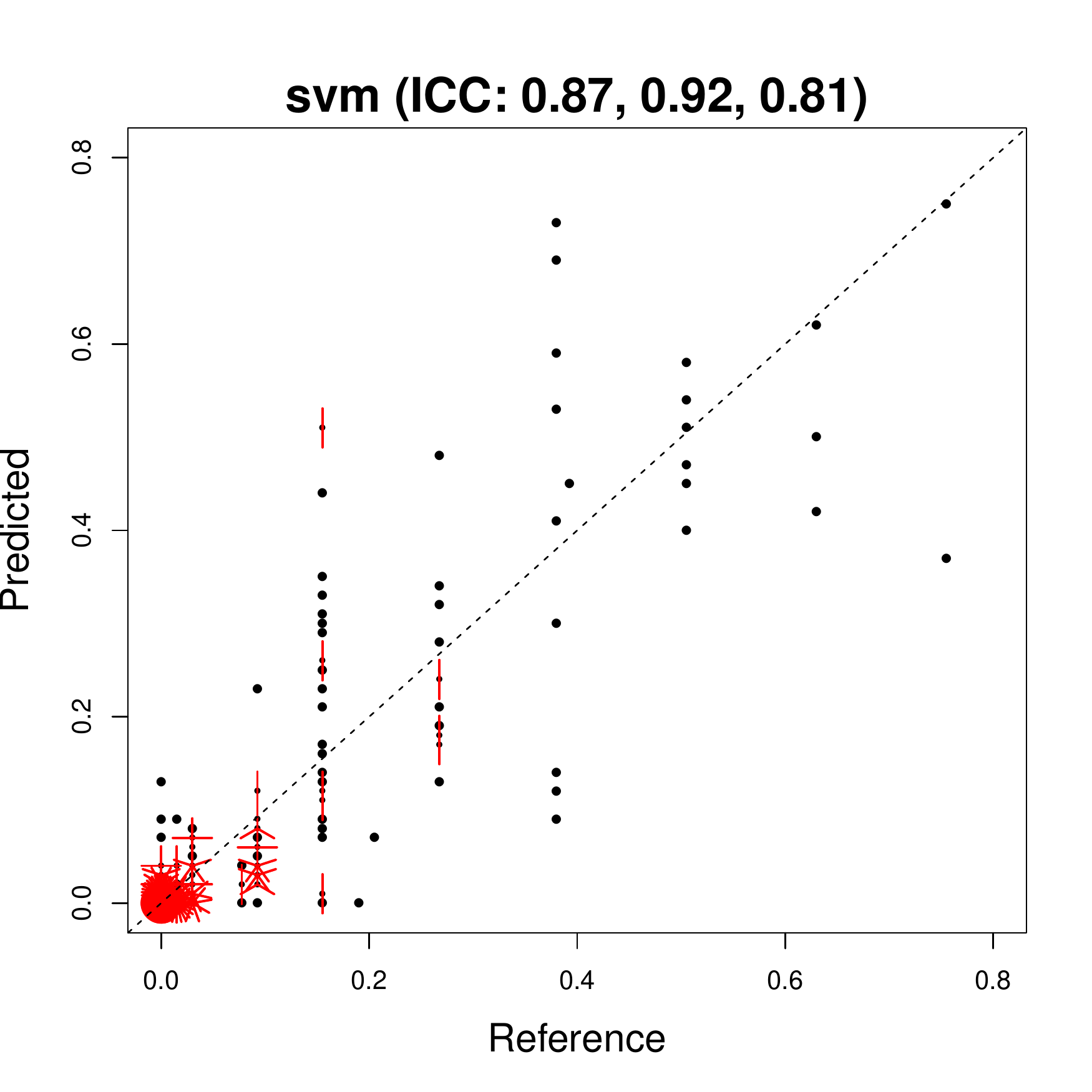}
  \includegraphics[width=0.313\textwidth,trim={55 00 30 25},clip=true]{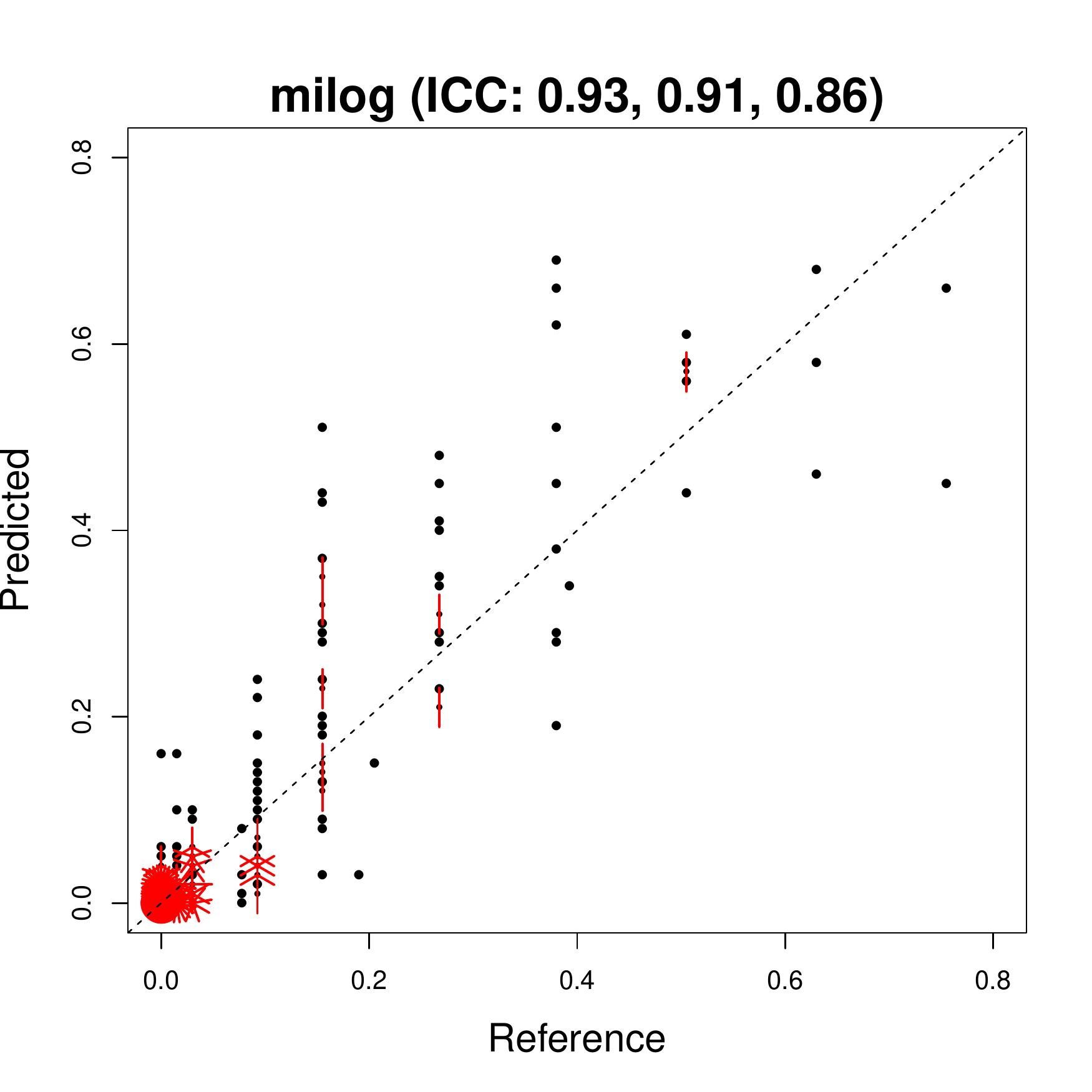}
  \includegraphics[width=0.354\textwidth,trim={00 00 30 25},clip=true]{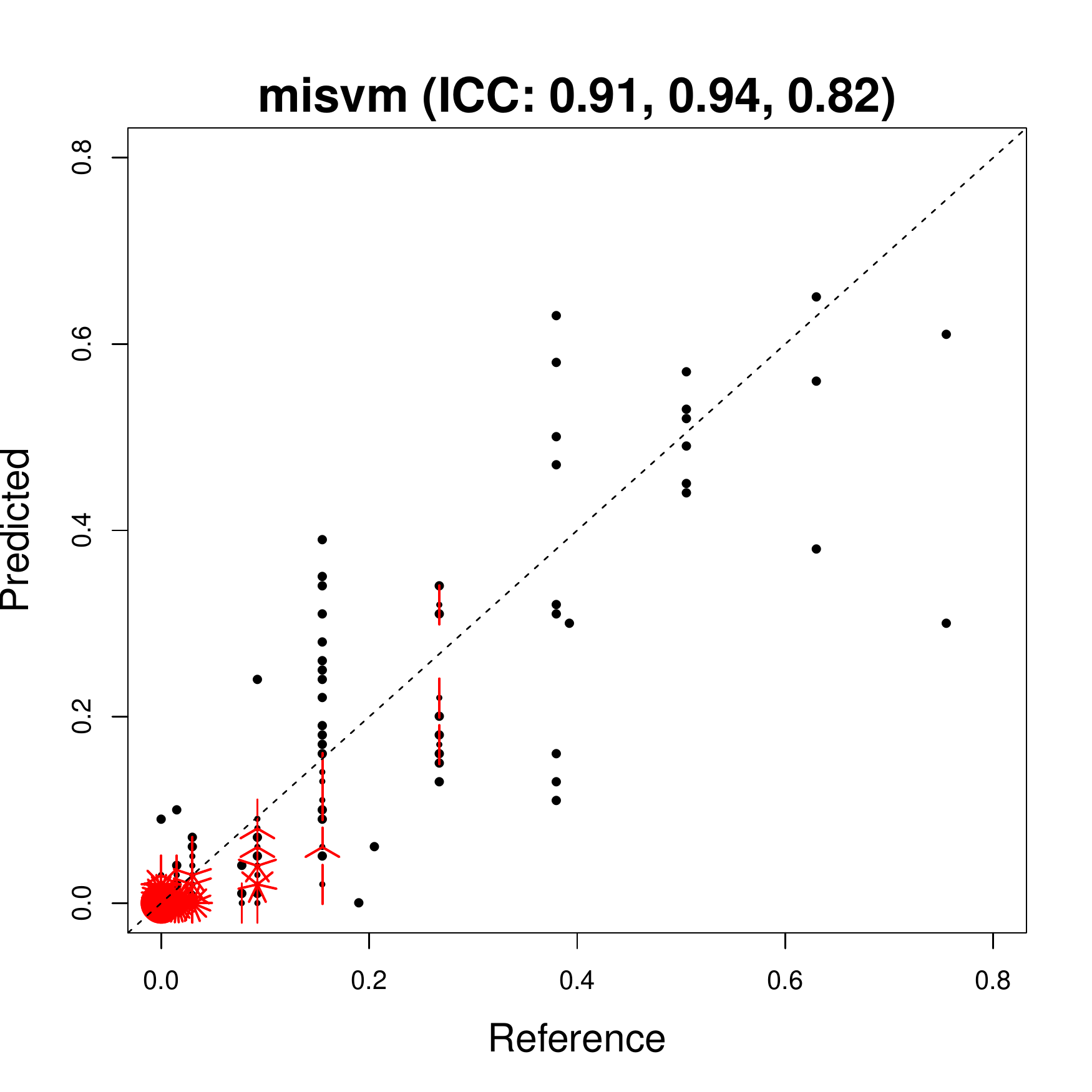}
  \includegraphics[width=0.313\textwidth,trim={55 00 30 25},clip=true]{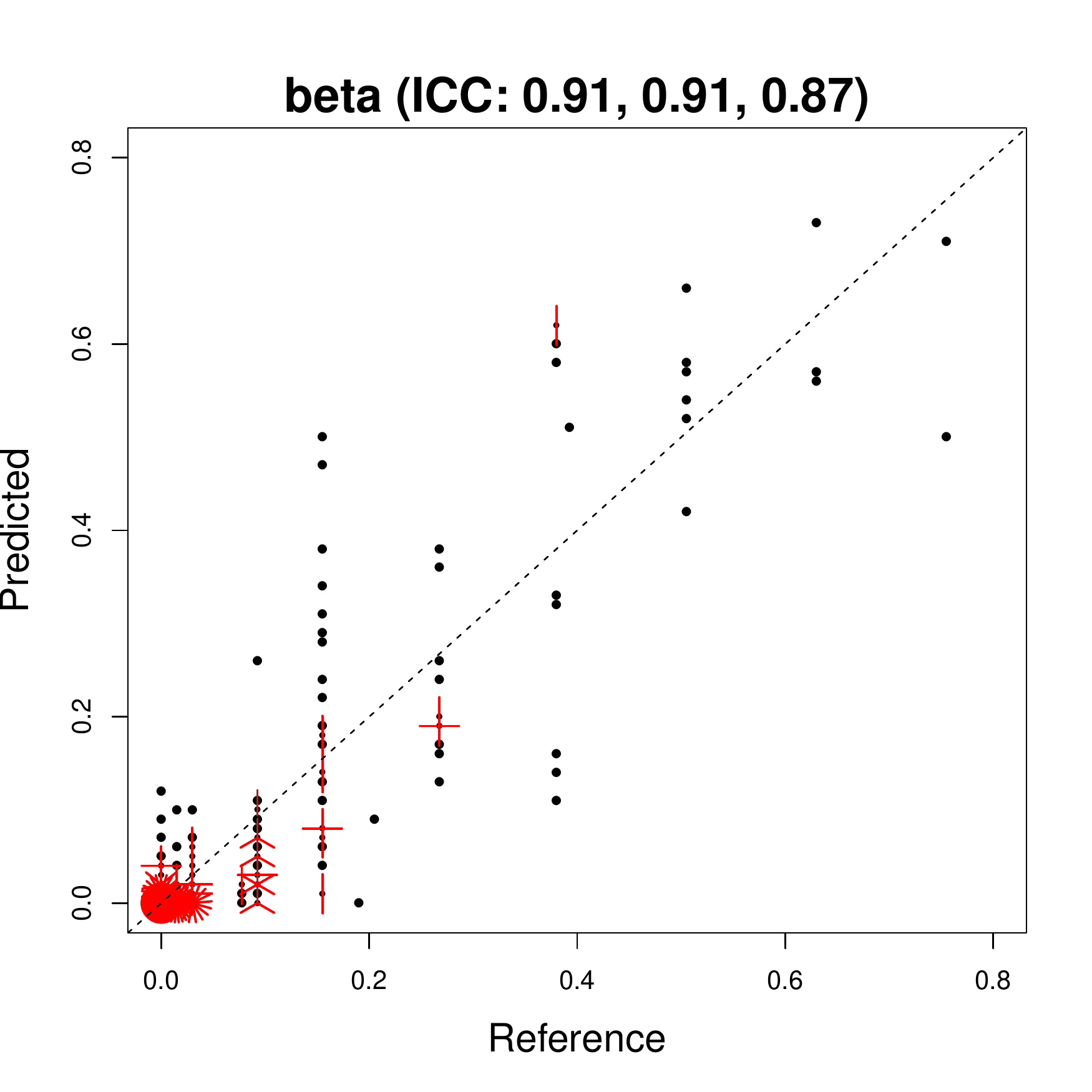}
  \includegraphics[width=0.313\textwidth,trim={55 00 30 25},clip=true]{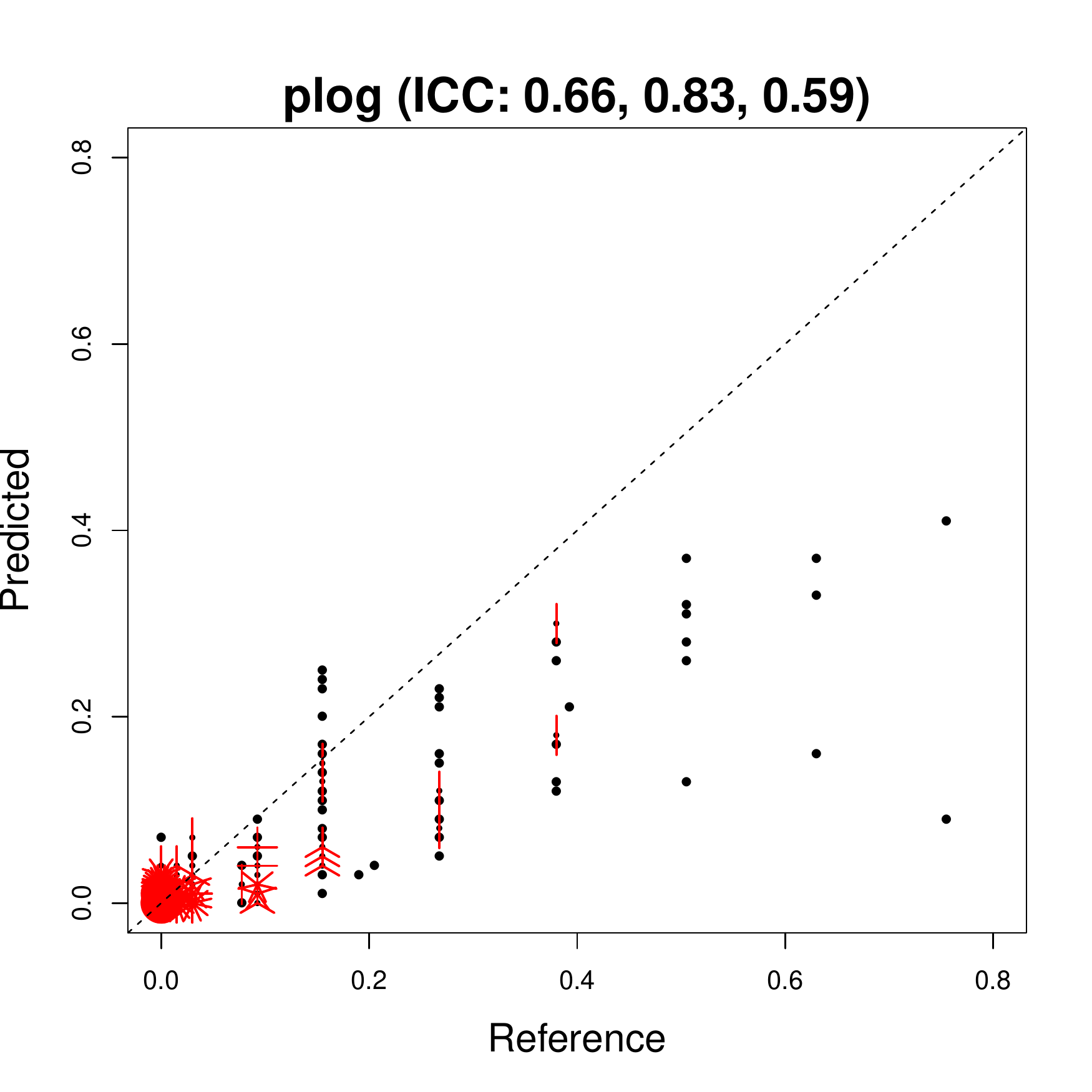}
  \includegraphics[width=0.354\textwidth,trim={00 00 30 25},clip=true]{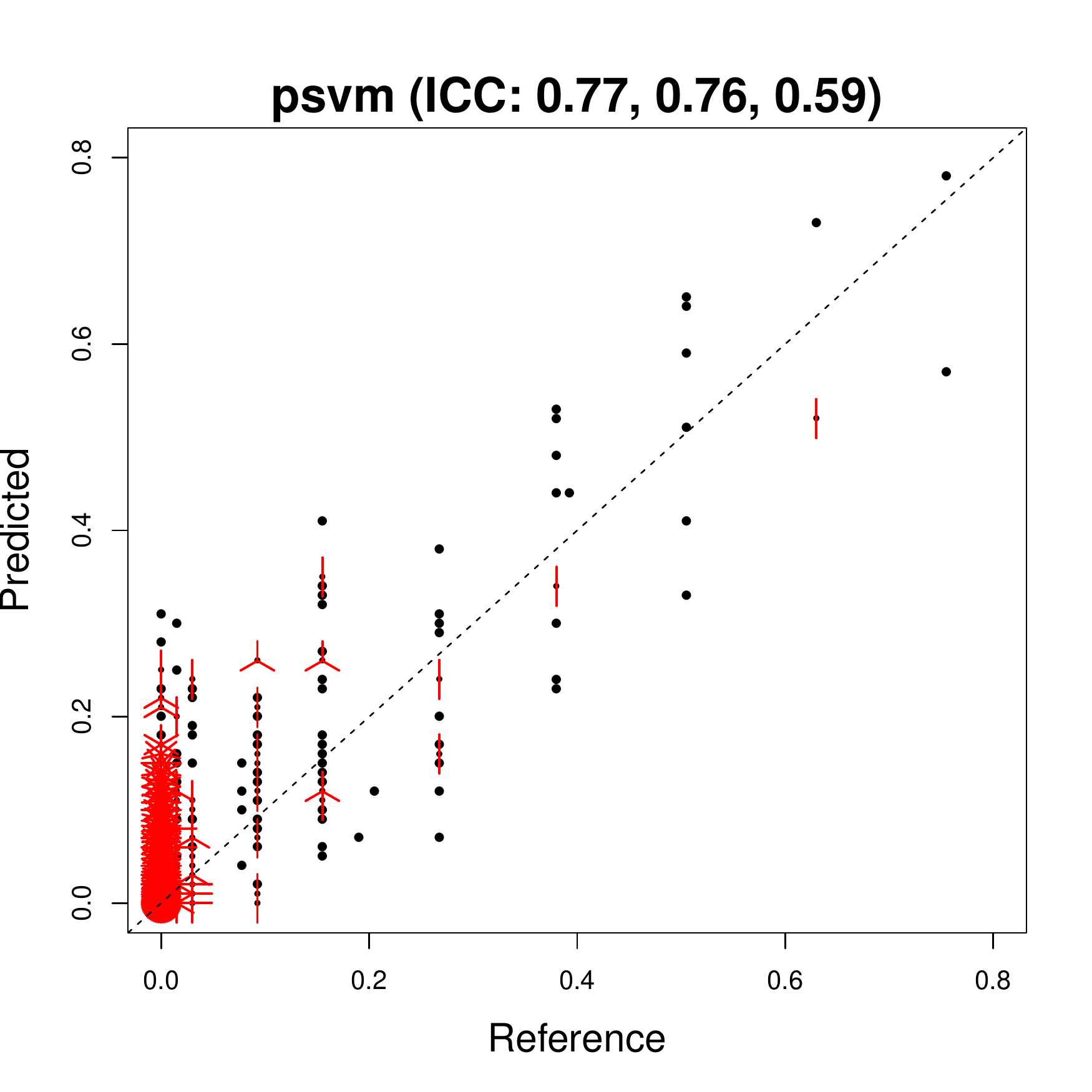}
  \includegraphics[width=0.313\textwidth,trim={55 00 30 25},clip=true]{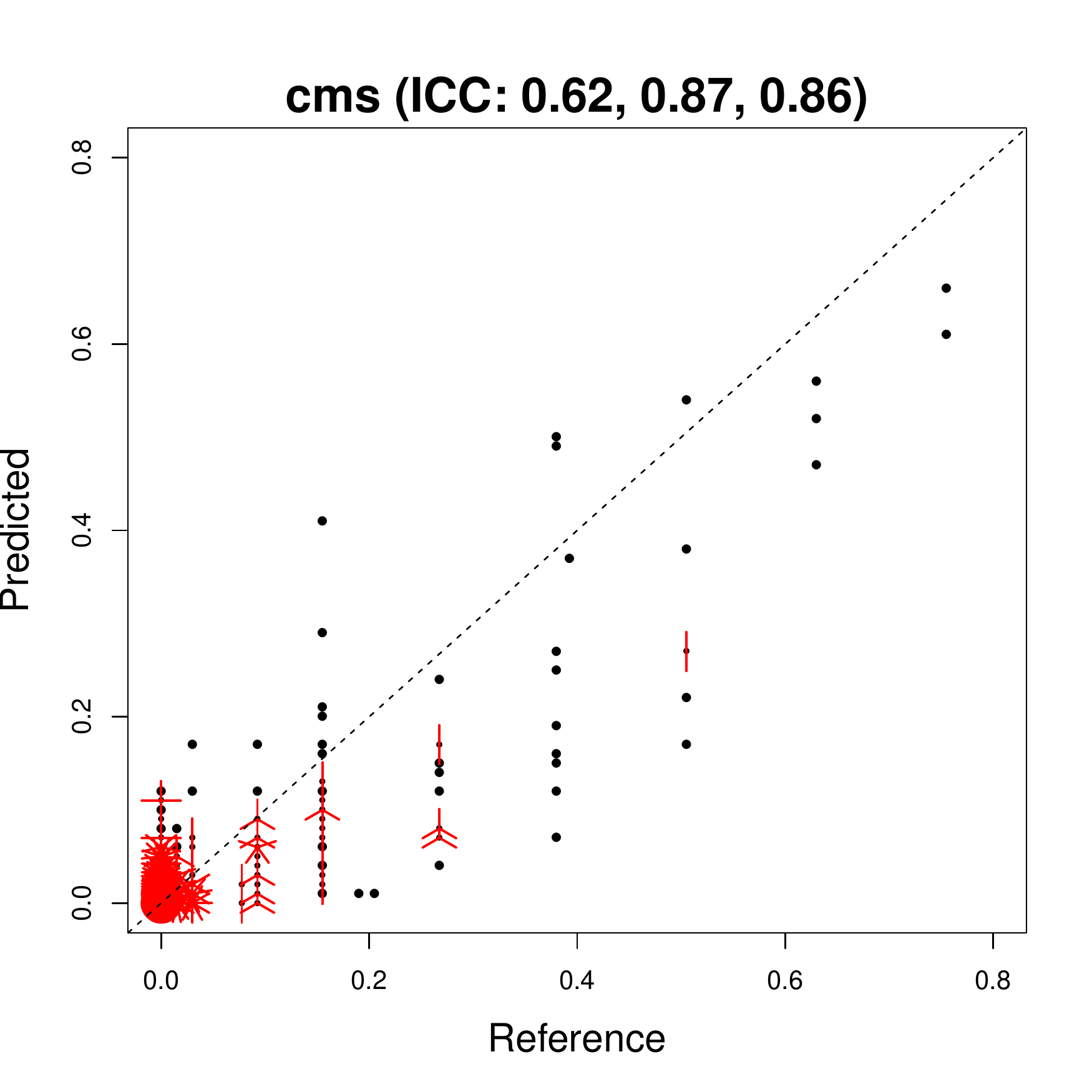}
  \includegraphics[width=0.313\textwidth,trim={55 00 30 25},clip=true]{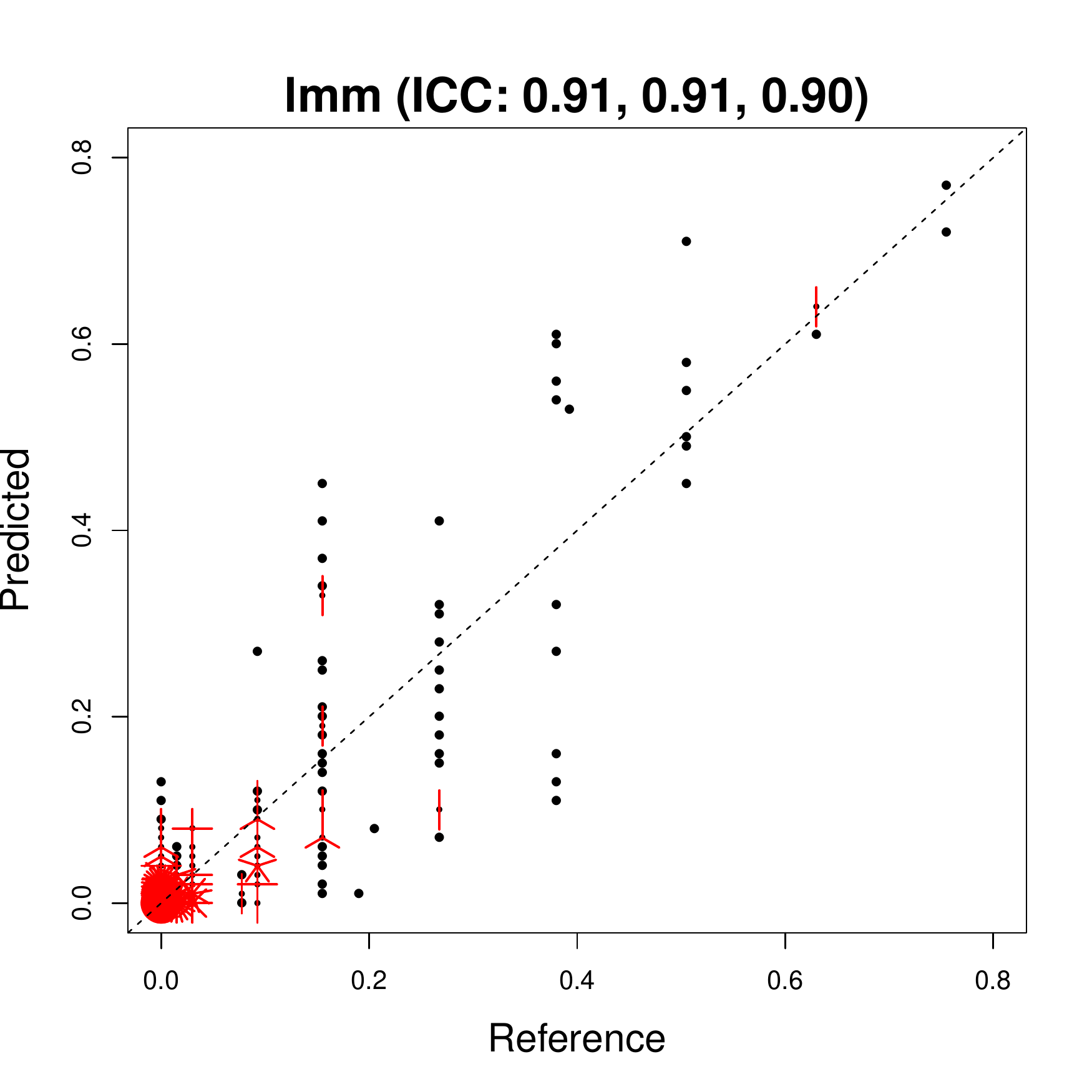}
  \caption{Correlation between predicted and reference extent of emphysema. The x-axis is reference extent and the y-axis is predicted extent. The amount of red ``petals'' at a coordinate indicates the amount of coincident points. Plot titles show ICC coefficients for each replication.}
  \label{fig:correlation-extent}
\end{figure*}

The prediction performance of the nine classifiers is illustrated with correlation plots in \figurename~\ref{fig:correlation-extent}. The numbers in the title of each plot are intra-class correlation coefficients (ICC, two-way model, agreement) for each replication. The average ICC coefficients over the three replications are shown in Table~\ref{tab:average-icc}. 
We see clear positive correlation between reference and predicted extent for all classifiers. It appears that plog and cms tend to underestimate extent, whereas psvm tends to overestimate for cases with low extent but seems to perform very well for larger extent.
Most classifiers show the largest variation for 15\% reference extent. For extent larger than 15\% we see very few cases with 0\% extent predicted. The ICC values across replications, also seen in \figurename~\ref{fig:correlation-extent}, illustrate that the performance of some classifiers varies a lot, with a difference of 0.25 in the worst case (cms). The most stable ICC performance is seen for lmm, which also has the highest average ICC.

\begin{table}[!t]
  \resizebox{\columnwidth}{!}{
    \begin{tabular}{cccc|ccccc}
      log  & svm  & milog & misvm & beta & plog & psvm & cms  & lmm \\
      \hline                             
      0.88 & 0.86 & 0.90  & 0.89  & 0.89 & 0.69 & 0.71 & 0.78 & 0.91
    \end{tabular}
  }
  \caption{Average ICC of of emphysema extent over the three replications. MIL on the left, LLP on the right.}
  \label{tab:average-icc}
\end{table}

\subsection{Replacing a rater}
\label{sec:replacing}
The ICC of predicted extent and average rater extent provides an overall measure of performance and a validation that the classifiers have learned what they are trained to do. We are also interested in how the classifiers compare against each rater on the original rater task, i.e assign one of six intervals of emphysema extent.
We converted predicted extent into the six extent intervals and calculated agreement with each rater. Agreement was calculated as described in Section~\ref{sec:measures} and is reported in Table~\ref{tab:agreement-clf-raters} as an average over raters and replications. The final column in Table~\ref{tab:agreement-clf-raters} provides inter-rater agreement averaged over replications.
We see that beta and misvm have the highest overall agreement (78 and 79), which is not far from the overall rater agreement of 83. The agreement pattern of misvm and beta also seem to match that of the raters to a large degree, with a large agreement on 0\% extent cases. It is interesting that psvm has the worst overall performance yet seems to outperform the other classifiers and raters for 51-75\% extent. However, we cannot rule out that this is just a random coincidence given the low prevalence of that class. Another interesting observation is that the best results relative to inter-rater agreement is seen for 6-25\% and 51-75\%, with four classifiers having better agreement scores than inter-rater agreement.

\begin{table}[!t]
  \resizebox{\columnwidth}{!}{
    \begin{tabular}{lccccccc}
      \,     & 0 & 1-5 & 6-25 & 26-50 & 51-75 & 76-100 & Overall\\
      \hline
      log    & 88  & 35    & 54     & 38      & 17      &  00       & 74\\
      svm    & 89  & 37    & 49     & 27      & 12      &  00       & 74\\
      milog  & 85  & 35    & 50     & 36      & 29      &  00       & 71\\
      misvm  & 91  & 39    & 58     & 36      & 31      &  00       & 79\\
      \hline 
      beta   & 91  & 35    & 54     & 24      & 47      &  00       & 78\\
      plog   & 72  & 26    & 57     & 45      & 00      &  00       & 58\\
      psvm   & 27  & 15    & 21     & 35      & 51      &  17       & 24\\
      cms    & 62  & 22    & 49     & 28      & 37      &  00       & 49\\
      lmm    & 81  & 31    & 49     & 30      & 44      &  17       & 66\\
      \hline 
      Rater  & 95  & 49    & 53     & 47      & 32      &  00       & 83
    \end{tabular}
  }
  \caption{Agreement percentages between classifiers and raters averaged over replications and raters. First four columns show MIL classifiers, next five columns show LLP classifiers, last column shows rater agreement.}
  \label{tab:agreement-clf-raters}
\end{table}

\subsection{Ranking classifiers}
\begin{figure}[!t]
  \centering
  \begin{tikzpicture}[%
    align=left,
    auto,
    block/.style={
      rectangle,
      anchor=base west,
      minimum height=2em
    },
    mymatrix2/.style={
      matrix of nodes, 
      nodes=block,
      column sep=-0.5em,
      row sep=1.3em,
    }
    ]
    \matrix[mymatrix2] (mx) {
      {\bf beta}  & misvm & log         &  svm       & milog &  {\bf lmm}  & {\bf plog} & {\bf cms}   & {\bf psvm} \\ 
      misvm       & log   & svm         & {\bf beta} & milog &  {\bf lmm}  & {\bf plog} & {\bf cms}   & {\bf psvm} \\
      milog       & misvm & {\bf beta}  & svm        & log   &  {\bf lmm}  & {\bf plog} & {\bf cms}   & {\bf psvm} \\
    };

    % D1
    \draw[black] ($(mx-1-1.north west)+(0, 0.2)$)    rectangle ($(mx-1-6.south east)+(-0.15,-0.25)$);
    \draw[black] ($(mx-1-5.north west)+(0, 0.1)$)    rectangle ($(mx-1-7.south east)+(-0.15,-0.09)$);
    \draw[black] ($(mx-1-7.north west)+(0.11, 0.0)$) rectangle ($(mx-1-8.south east)+(-0.15, 0.02)$);
    \draw[black] ($(mx-1-8.north west)+(0.1,-0.1)$)  rectangle ($(mx-1-9.south east)+(-0.15, 0.12)$);

    % D2
   \draw[black] ($(mx-2-1.north west)+(0, 0.2)$) rectangle ($(mx-2-6.south east)+(-0.15,-0.2)$);
   \draw[black] ($(mx-2-2.north west)+(0, 0.1)$) rectangle ($(mx-2-7.south east)+(-0.15,-0.05)$);

    % D3
   \draw[black] ($(mx-3-1.north west)+(0, 0.2)$) rectangle ($(mx-3-7.south east)+(-0.15,-0.2)$);
   \draw[black] ($(mx-3-3.north west)+(0.05, 0.05)$) rectangle ($(mx-3-8.south east)+(-0.15,-0.05)$);

  \end{tikzpicture}
   \caption{Grouping of classifiers based on difference in extent prediction performance as decided by the Nemenyi test. Classifiers in the same box are not significantly different ($\alpha=0.05$). Columns are sorted by mean rank over all test samples in descending order. {\bf Bold} typeface indicates LLP methods.}
   \label{fig:extent-significance-grouping}
\end{figure}
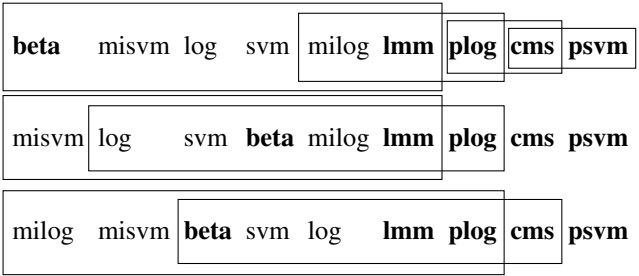

We use Friedmans and Nemenyis test for comparing classifiers as suggested in \cite{demsar2006statistical}. We test the hypothesis {\em $H_0:$ All classifiers are equal} using Friedmans test and significance level $\alpha = 0.05$. This test is based on the rank of the classifiers for each sample prediction. We use the absolute distance from predicted extent to reference extent to assign ranks. In all three replications we get $p < 0.001$ for the Friedman test and reject the hypothesis that all classifiers have equal performance. We then test the pairwise hypothesis {\em $H_0:$ The classifiers are equal} for all pairs of classifiers using the Nemenyi test. The results of the Nemenyi tests are summarized in \figurename~\ref{fig:extent-significance-grouping}. Columns are sorted by average ranks and $H_0$ is rejected for classifiers that are not in the same box. We see that the LLP methods plog, cms and psvm are consistently ranked low, confirming the low ICC in Table~\ref{tab:average-icc} and the low overall agreement in Table~\ref{tab:agreement-clf-raters}. Even though lmm is never significantly different from the best classifier, it is consistently ranked low. We also saw in Table~\ref{tab:agreement-clf-raters} that lmm had low overall agreement with raters yet achieved the best average ICC. It is also interesting that misvm is consistently ranked in the top-2.

\subsubsection{Label stability}
\label{sec:extent-label-stability}
We investigate label stability under changes in training data by predicting all test data with the trained models from each replication. For each classifier we got three sets of predictions of 60,000 instances and 600 bags. We converted bag predictions to the six extent intervals and measured agreement between replications for predicted bag and instance labels for each classifier. 
The stability results are summarized in Table~\ref{tab:extent-label-stability}. For bag labels, most classifiers have best agreement on 0\% extent followed by 6-25\%. Overall, beta and misvm are the most stable classifiers for both bag and instance labels, whereas milog is the most stable classifier on 6-25\%, 26-50\% and 51-75\%. The missing scores for 51-75\% and 76-100\% are because there are no predictions of these classes in any of the replications. The inter-rater agreement on bag labels is included in the last row of Table~\ref{tab:extent-label-stability}. We see that misvm and beta always have equal or better agreement than the raters, and most methods have better agreement than raters on all non-zero extent scores.

\begin{table*}[!t]
  \centering
  %\resizebox{\textwidth}{!}{
    \begin{tabular}{l|ccccccc|cc}
      \multicolumn{1}{c}{}   & \multicolumn{7}{c}{Bag}                          & \multicolumn{2}{c}{Instance}\\
      \,    & 0\%    & 1-5\%  & 6-25\% & 26-50\% & 51-75\% & 76-100\% & Overall & E     & NE  \\
      \hline                                                                 
      log   &   89   &   60   &   68   &   58    &   57    & --       &   79    &   35  &   97\\
      svm   &   89   &   57   &   60   &   51    &   58    & --       &   78    &   44  &   97\\
      milog &   84   &   60   &\bf{81} &\bf{82}  &\bf{77}  & --       &   78    &   44  &   96\\
      misvm &   95   &   70   &   77   &   70    &   74    & --       &   88    &   52  &\bf{98}\\
      \hline
      beta  &\bf{96} &\bf{71} &   72   &   58    &   62    &\bf{50}   &\bf{89}  &\bf{56}&\bf{98}\\
      plog  &   65   &   54   &   73   &   78    & --      & --       &   61    &   07  &   95\\
      psvm  &   24   &   40   &   47   &   47    &   60    & 0        &   41    &   12  &   82\\
      cms   &   65   &   60   &   59   &   45    &   12    & --       &   61    &   05  &   93\\
      lmm   &   82   &   60   &   68   &   58    &   67    &   17     &   73    &   43  &   97\\
      \hline
      Rater &   95   &   49   &   53   &   47    &   32    &   0      &   83    &   --  &   --
    \end{tabular}
  %}
  \caption{Label stability. Agreement percentages between predictions from each replication. Instance columns are for binary instance predictions ({\bf E}mphysema / {\bf N}o {\bf E}mphysema). A dash (--) indicates no predictions in that category.}
  \label{tab:extent-label-stability}
\end{table*}

\section{Discussion \& Conclusion}
\label{sec:discussion}
We have focused on comparing MIL methods, which have previously shown promising results for COPD and emphysema detection, with LLP methods that can learn directly from proportion labels. While end-to-end learning using CNNs have shown promising results for medical imaging tasks, and have just recently been used for emphysema quantification \cite{bortsova2018deep}, we decided to use classic scale space features to focus on the aspects of learning from binary versus proportion labels, and to establish performance of classic feature engineering approaches.

Using the average rater as reference, the best classifiers achieve ICC coefficients around 0.9. Average overall agreement between the best classifiers and each rater on six emphysema extent intervals is close to the inter-rater agreement (78-79\% vs 83\%). For some extent intervals the classifiers are better than the inter-rater agreement. These results show that that the presented approach to automatic emphysema extent prediction is viable and could be useful for routine assessment of emphysema extent.

The four best performing classifiers, beta, misvm, milog and lmm, have very similar ICC coefficients, with lmm being slightly more consistent across replications. However, beta and misvm show superior overall prediction of extent intervals with a much better discrimination of CT scans without visible emphysema compared to milog and lmm. Overall stability of beta and misvm is also superior to milog and lmm, although milog shows more stable predictions for the lower prevalence extent intervals 6-25\%, 26-50\% and 51-75\%.
Learning from scores indicating emphysema extent did not appear to be advantageous for extent prediction compared to learning based on emphysema presence alone. The MIL classifiers, misvm and milog, and the LLP classifiers, beta and lmm, show comparable performance. 

One possible explanation for the lack of improved performance when training on extent labels, is that the extent labels are too noisy, as the relatively large disagreement between observers suggests. Obtaining more accurate and precise extent labels is costly and it is not clear if it is possible to improve the label quality significantly. 
In this work we have combined the emphysema estimates of two raters by simple averaging of point estimates. In \cite{orting2016quantifying} we showed that performance of the cms classifier improved when learning from labels incorporating rater uncertainty over learning from averaged point estimates. The approach used in \cite{orting2016quantifying} is not directly applicable to the other methods used here and we have used point estimates to keep the comparison fair. 
Recent work on classification of retinal images with a CNN-based method \cite{guan2017who} show that modeling individual raters can improve performance over simple averaging of multiple raters. Although more than 30 raters were used in \cite{guan2017who} it is possible that a more complex model of rater annotations could also improve performance when only two raters are used.

Another possible factor is that the model of proportion labels is too simple to exploit the additional information in the labels. The results in \cite{bortsova2018deep} indicate that learning from proportion labels can help more complex models based on CNNs to converge faster and to a better optima than learning from binary labels. A possible explanation for this is that explicitly modeling proportion labels has a regularizing effect on the feature learning part of CNNs, which would also explain why we do not see improved performance for LLP methods when features are fixed.

We considered three strategies for learning from bag labels, the simple strategy, the relabeling strategy and the mean strategy. For the MIL classifiers it appears that the relabeling strategy is best, whereas for the LLP classifiers it appears that the simple and mean strategies are best. One reason the simple strategy works better for LLP than for MIL could be that a proportion instance label as used in simple LLP is interpreted as a probability of emphysema in the patch, whereas the binary instance labels as used in simple  MIL are interpreted as the probability the patch came from a CT scan with emphysema. In this sense, the proportion instance labels match the intended objective, predicting the proportion of patches with emphysema, much better than the binary instance labels.

A limitation of this study is that we have only trained and validated the classifiers on the upper right region of the lung. Due to the lower prevalence and agreement of visual scoring in the remaining five regions, we expect some decrease in extent prediction accuracy for these regions, similar to what was observed in \cite{orting2018detecting} for regional emphysema detection. Investigating the performance over all regions should be considered in future work. However, the results in \cite{orting2018detecting} show that a simple MIL classifier trained on subject-level presence/absence labels can provide the same performance as a classifier trained on region-level presence/absence labels. In light of the results here, this suggests that a MIL classifier, such as misvm,  could provide accurate regional emphysema extent estimates even when trained only on subject-level presence of emphysema.

In conclusion, the best performing classifiers have close to human-level performance and are promising candidates for automatic quantification of emphysema extent. Furthermore, MIL classifiers having access to only emphysema presence labels perform just as well as LLP classifiers with access to emphysema extent labels. Reducing the labeling task from estimating emphysema extent to indicating presence, reduces the cost of training and makes it more feasible to implement in new settings.

\section{References}
\bibliography{IEEEabrv,References}

\clearpage
\appendices
\section{Emphysema}
\label{app:emphysema}
\begin{figure*}[!t]
  \centering
  \includegraphics[width=0.32\textwidth]{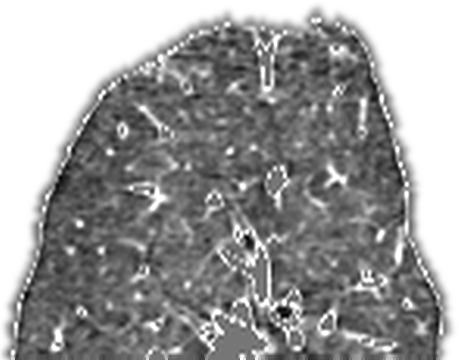}%
  \includegraphics[width=0.32\textwidth]{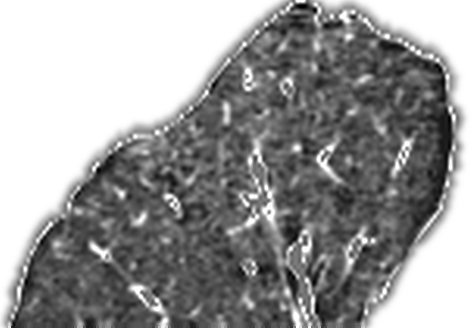}%
  \includegraphics[width=0.32\textwidth]{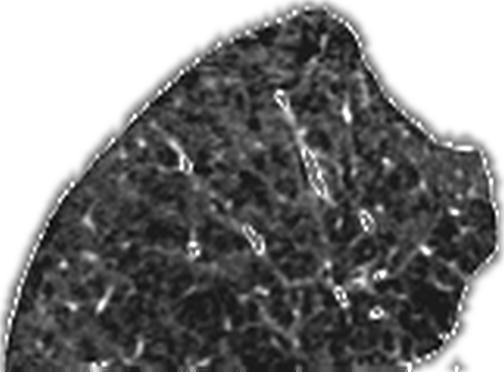}
  \caption{Example slices. From left, visually assessed emphysema extent is 0\%, 6-25\% and 51-75\%. Window level -780HU, window width 560HU.}
  \label{fig:example-slices}
\end{figure*}

\figurename~\ref{fig:example-slices} shows slices from the upper right region of three CT scans. Background and airways have been masked. The left image is assessed as having no visible emphysema extent. The center image as having 6-25\% and the right image as having 51-75\% emphysema extent. For the center image, emphysema is predominately visible at the boundary of the lung, whereas it is distributed throughout the region in the right image.

\section{Methods}
\label{app:methods}
\subsection{Notation}
Let $\mathcal{X}$ be an instance space, $\mathcal{Y}$ an instance label space, $\mathcal{Z}$ a bag label space and $\mathbf{b} = (\mathbf{x} \subseteq \mathcal{X}, z \in \mathcal{Z})$ a labeled bag of instances. We use superscripts to refer to the label ($\mathbf{b}^z$), instances ($\mathbf{b}^{\mathbf{x}}$) and instance labels ($\mathbf{b}^{\mathbf{y}}$) associated with a bag $\mathbf{b}$. For a set of $m$ bags $\mathbf{B} = \{\mathbf{b}_1, \mathbf{b}_2, \dots, \mathbf{b}_m\}$, $\mathbf{b}^{\mathbf{x}}_i$ are the instances in the $i$'th bag and $\mathbf{b}^{\mathbf{x}}_{ij}$ is the $j$'th instance in the $i$'th bag. For the set of all instances we use $\mathbf{X} = \cup_{i=1}^m \mathbf{b}^{\mathbf{x}}_i$, for all instance labels we use $\mathbf{Y} = \cup_{i=1}^m \mathbf{b}^{\mathbf{y}}_i$ and for all bag labels we use $\mathbf{Z} = \cup_{i=1}^m \{\mathbf{b}^z_i\}$.

\subsection{$mi$-logistic}
The bag learning problem for $mi$-logistic is a constrained optimization problem over model weights and unknown instance labels
\begin{align}
  \label{eq:mi-logistic-bag-learning}
  &\max\limits_{\mathbf{w},\mathbf{Y}} \prod\limits_{i,j} p(\mathbf{b}^\mathbf{y}_{i,j} \,|\, \mathbf{b}^\mathbf{x}_{i,j}, \mathbf{w})\\
  \mathrm{s.t.} &\forall i : \Theta_{\max}(\mathbf{b}^\mathbf{y}_i) = \mathbf{b}^z_i \in \{0,1\}.
\end{align}
We use the heuristic for solving the $mi$-SVM problem from \cite{andrews2003support}. Initially, fix instance labels by setting them to bag labels, $\mathbf{b}^\mathbf{y}_{i,j} = \mathbf{b}^z_i \forall i,j$. For fixed instance labels (\ref{eq:mi-logistic-bag-learning}) reduces to standard logistic regression. Let $h(\cdot) = \sigma(\mathbf{w}^T\cdot)$ denote the fitted model. Instance labels are predicted as 
\begin{equation}
\tilde{\mathbf{b}}^\mathbf{y}_{i,j} = \mathds{1}\{ h(\mathbf{b}_{i,j}^\mathbf{x}) > 0.5\}
\end{equation}
and bag labels are predicted as 
\begin{equation}
\tilde{\mathbf{b}}^z_i = \Theta_{\max}(\tilde{\mathbf{b}}^\mathbf{y}_i).
\end{equation}
Instance labels are then updated according to
\begin{equation}
  \label{eq:mi-svm-update}
  \mathbf{b}^\mathbf{y}_{i,j} = \threepartdef{0}     {\mathbf{b}^z_i = 0}
                                           {1}     {\mathbf{b}^z_i = 1. \,\tilde{\mathbf{b}}^z_i = 0.\, h(\mathbf{b}^\mathbf{x}_{i,j}) >  h(\mathbf{b}^{\mathbf{x}}_{i,k}) \forall k \ne j}
                                           {\tilde{\mathbf{b}}^{\mathbf{y}}_{i,j}} {\text{otherwise}}
\end{equation}
The first clause ensures that instances from negative bags are always labeled negative. The second clause ensures that a positive bag predicted as negative will always have one positive instance by labeling the ``most'' positive instance as positive, and the third clause ensures all other instances in positive bags are relabeled to match the predicted class. 

\subsection{$\propto$-logistic}
\label{sec:propto-logistic}
The $\propto$-logistic model can be derived by considering the joint probability over instances $\mathbf{X}$, bag labels $\mathbf{Z}$ and instance labels $\mathbf{Y}$
\begin{align}
  P(\mathbf{Y},\mathbf{X},\mathbf{Z}) &= P(\mathbf{Z} | \mathbf{Y},\mathbf{X}) P(\mathbf{Y},\mathbf{X}) \\
           &= P(\mathbf{Z} | \mathbf{Y})   P(\mathbf{Y},\mathbf{X}) & \mathbf{Z} \independent \mathbf{X} \,|\, \mathbf{Y} \\
           &\propto P(\mathbf{Z}|\mathbf{Y}) P(\mathbf{Y}|\mathbf{X}) & P(\mathbf{X}) = \mathrm{Constant}\label{eq:propto-logistic}
\end{align}
We use a logistic model for instance labels and a binomial model for bag labels.
\begin{align}
  P(\mathbf{Y} \,|\, \mathbf{X}) =& \prod\limits_{i,j} P(\mathbf{b}^{\mathbf{y}}_{i,j}|\mathbf{b}^{\mathbf{x}}_{i,j},\mathbf{w}) \\
      =& \prod\limits_{i,j} \sigma(\mathbf{w}^T\mathbf{b}^{\mathbf{x}}_{i,j})^{\mathbf{b}^{\mathbf{y}}_{i,j}} (1 - \sigma(\mathbf{w}^T\mathbf{b}^{\mathbf{x}}_{i,j}))^{1 - \mathbf{b}^{\mathbf{y}}_{i,j}} \label{eq:logistic}  \\
  P(\mathbf{Z} \,|\, \mathbf{Y} ) =& \prod\limits_i P(\mathbf{b}^z_i | \mathbf{b}^{\mathbf{y}}_i) =\\
                                 \prod\limits_i \binom{|\mathbf{b}_i|}{|\mathbf{b}_i| \mathbf{b}^z_i}& \Theta_{\mathrm{mean}}(\mathbf{b}^{\mathbf{y}}_i)^{|\mathbf{b}_i| \mathbf{b}^z_i} (1 - \Theta_{\mathrm{mean}}(\mathbf{b}^{\mathbf{y}}_i)^{|\mathbf{b}_i| - {|\mathbf{b}_i| \mathbf{b}^z_i}} \label{eq:binom}
\end{align}
substituting into (\ref{eq:propto-logistic}) gives us
\begin{align}
  P(\mathbf{Z}|\mathbf{Y}) P(\mathbf{Y}|\mathbf{X}) &= \prod\limits_i P(\mathbf{b}^z_i | \mathbf{b}^{\mathbf{y}}_i) \prod\limits_{j} P(\mathbf{b}^{\mathbf{y}}_{i,j}|\mathbf{b}^{\mathbf{x}}_{i,j},\mathbf{w})
\label{eq:propto-logistic-obj}
\end{align}
We want to find the $\mathbf{Y}$ and $\mathbf{w}$ that maximize (\ref{eq:propto-logistic-obj})
\begin{align}
  \arg\max\limits_{\mathbf{Y},\mathbf{w}} \prod\limits_i P(\mathbf{b}^z_i | \mathbf{b}^{\mathbf{y}}_i) \prod\limits_{j} P(\mathbf{b}^{\mathbf{y}}_{i,j} \,|\, \mathbf{b}^{\mathbf{x}}_{i,j},\mathbf{w})
\end{align}
We do this by fixing $\mathbf{Y}$ and $\mathbf{w}$ iteratively. For fixed $\mathbf{Y}$ we get standard logistic regression. For fixed $\mathbf{w}$ we can optimize over each bag individually
\begin{align}
  \arg\max\limits_{\mathbf{b}^{\mathbf{y}}_i} P(\mathbf{b}^{z}_i \,|\, \mathbf{b}^{\mathbf{y}}_i ) \prod_{j=1} P(\mathbf{b}^{\mathbf{y}}_{i,j} \,|\, \mathbf{b}^{\mathbf{x}}_{i,j},\mathbf{w})  \label{eq:fixed-w}.
\end{align}
This can be done with the same greedy method used for $\propto$-SVM in \cite{yu2013proptosvm}.

\section{Parameters}
\label{app:parameters}
All classifiers provide probability estimates of instance labels and a classifier-specific instance threshold was fitted on the training data by trying all thresholds in the range $[0, 0.01, 0.02, \dots, 0.99, 1]$. Fitted thresholds are reported in Table~\ref{tab:thresholds}. There is a large variation in fitted instance thresholds across classifiers, and for some classifiers there is a large variation across replications. Variation across replications is an indication that the classifier has learned substantially different decision rules for each replication. Variation between classifiers could just be a scaling issue, but is at least an indication that interpreting instance predictions as probability estimates is problematic.
\begin{table}[!t]
  \centering
  %\resizebox{0.6\columnwidth}{!}{
  \begin{tabular}{cccc}
    Classifier & D1   & D2   & D3\\
    \hline
    log        & 0.78 & 0.79 & 0.60\\
    beta       & 0.09 & 0.09 & 0.09\\
    svm        & 0.77 & 0.76 & 0.70\\
    misvm      & 0.85 & 0.94 & 0.78\\
    milog      & 0.99 & 0.99 & 0.99\\
    psvm       & 0.97 & 0.99 & 0.14\\
    plog       & 0.01 & 0.01 & 0.01\\
    cms        & 0.86 & 0.68 & 0.99\\
    lmm        & 0.75 & 0.62 & 0.73
  \end{tabular}
%}
  \caption{Fitted instance thresholds for each classifier and replication}
  \label{tab:thresholds}
\end{table}

\subsection{beta}
\label{sec:betas}
The implementation of beta regression requires uncorrelated features and we used the PCA algorithm to decorrelate features. We tried dimensionality reduction (only keep principal components with standard deviation $\ge 1$). We tried two optimization methods, maximum likelihood estimation (ML) and bias correction (BC).

Fitted parameters
\begin{description}
\item[beta]
  \begin{description}
  \item[D1] no dimensionality reduction, ML 
  \item[D2] no dimensionality reduction, BC
  \item[D3] no dimensionality reduction, ML
  \end{description}
\end{description}

\subsection{svm, misvm, psvm}
\label{sec:svms}
For all three classifiers we tried both linear and RBF kernels. In both cases we tried $C \in \{0.1, 1, 10, 100\}$. For psvm we tried $C_2 \in \{1, 10, 100, 1000\}$. For the RBF kernels we tried $\gamma \in \{0.1, 1\}$. We used Platt calibration \cite{lin2007note} to obtain probability estimates from all three SVMs. 

Fitted parameters
\begin{description}
\item[svm]
  \begin{description}
  \item[D1] linear kernel, $C = 1$
  \item[D2] linear kernel, $C = 0.1$
  \item[D3] linear kernel, $C = 10$
  \end{description}
\item[misvm] 
  \begin{description}
  \item[D1] linear kernel, $C = 0.1$
  \item[D2] linear kernel, $C = 0.1$
  \item[D3] linear kernel, $C = 0.1$
  \end{description}
\item[psvm]
  \begin{description}
  \item[D1] rbf kernel, $C = 1  , C_2 = 1, \gamma = 0.1$ 
  \item[D2] rbf kernel, $C = 0.1, C_2 = 1, \gamma = 1$
  \item[D3] rbf kernel, $C = 1  , C_2 = 1, \gamma = 0.1$
  \end{description} 
\end{description}

\subsection{log, milog, plog}
\label{sec:logs}
We tried dimensionality reduction using PCA for log. We did not use dimensionality reduction for milog and plog. We ran milog and plog until convergence of instance labels or for 20 iterations, whichever came first.

Fitted parameters
\begin{description}
\item[log]
  \begin{description}
  \item[D1] no dimensionality reduction 
  \item[D2] no dimensionality reduction
  \item[D3] dimensionality reduction
  \end{description}
\end{description}

\subsection{cms}
We used the following fixed parameters, branching = 2, number of $k$-means iterations = 25, maximum iterations of CMA-ES = 1000, $\lambda = 13$. We tried number of clusters $k \in \{10, 20, 30, 40, 50, 60, 70, 80, 90, 100\}$.

Fitted parameters
\begin{description}
\item[cms]
  \begin{description}
  \item[D1] $k = 70$
  \item[D2] $k = 50$
  \item[D3] $k = 100$
  \end{description}
\end{description}

\subsection{lmm}
We tried 
\begin{align*}
  \lambda &\in \{0, 1, 10, 100\}\\
  \gamma &\in \{0.00001, 0.0001, 0.001, 0.01, 0.1, 1\}\\
  &\sigma \in \{0.001,0.01,0.1,0.125,0.25, 0.5, 1.0\}
\end{align*}

Fitted parameters
\begin{description}
\item[lmm]
  \begin{description}
  \item[D1] $\lambda = 1, \gamma = 0.01, \sigma = 0.1$
  \item[D2] $\lambda = 1, \gamma = 0.01, \sigma = 0.1$
  \item[D3] $\lambda = 10, \gamma = 0.00001, \sigma = 0.25$
  \end{description}
\end{description}

\end{document}